\documentclass[letterpaper, 10 pt, conference]{ieeeconf}  % Comment this line out if you need a4paper

\IEEEoverridecommandlockouts                              % This command is only needed if 
\usepackage{graphics} % for pdf, bitmapped graphics files
\usepackage{epsfig} % for postscript graphics files
\usepackage{times} % assumes new font selection scheme installed
\usepackage{amsmath} % assumes amsmath package installed
\usepackage{amssymb}  % assumes amsmath package installed
\usepackage{bbding}
\usepackage{multirow}

\usepackage{gensymb}
\usepackage{cite}
\usepackage{algorithmic}
\usepackage{textcomp}
\usepackage{scrextend}
\usepackage{changepage}
\usepackage{multirow}
\usepackage{booktabs}
\usepackage{mathtools}
\usepackage{xcolor}
\usepackage{float}
\usepackage{makecell}

\usepackage{hyperref}
\usepackage{amsfonts}

\title{\LARGE \bf
RAIL: Robot Affordance Imagination with
Large Language Models
}

\author{Ceng Zhang$^{1*}$, Xin Meng$^{1*}$, Dongchen Qi$^{2}$, Gregory S. Chirikjian$^{1,2}$% <-this % stops a space
\thanks{* denotes equal contribution.}% <-this % stops a space
\thanks{This work was supported by NUS Startup grants A-0009059-02-00, A-0009059-03-00, CDE Board account E-465-00-0009-01, and National Research Foundation, Singapore, under its Medium Sized Centre Programme - Centre for Advanced Robotics Technology Innovation (CARTIN), sub award A-0009428-08-00.}
\thanks{$^{1}$ Ceng Zhang, Xin Meng and Gregory S. Chirikjian are with the Department of Mechanical Engineering, National University of Singapore, Singapore. \texttt{\{tmy\_zc, mengxin, mpegre\}@nus.edu.sg}} %
\thanks{$^{2}$ Dongchen Qi and Gregory S. Chirikjian are with the Department of Mechanical Engineering, University of Delaware, Newark, DE 19716, USA. \texttt{\{dcqi, gchirik\}@udel.edu}}
}

\begin{document}

\maketitle
\thispagestyle{empty}
\pagestyle{empty}

%%%%%%%%%%%%%%%%%%%%%%%%%%%%%%%%%%%%%%%%%%%%%%%%%%%%%%%%%%%%%%%%%%%%%%%%%%%%%%%%
\begin{abstract}
This paper introduces an automatic affordance reasoning paradigm tailored to minimal semantic inputs, addressing the critical challenges of classifying and manipulating unseen classes of objects in household settings.
Inspired by human cognitive processes, our method integrates generative language models and physics-based simulators to foster analytical thinking and creative imagination of novel affordances.
Structured with a hierarchical framework consisting of analysis, imagination, and evaluation, our system “analyzes” the requested affordance names into interaction-based definitions, “imagines” the virtual scenarios, and “evaluates” the object affordance.
If an object is recognized as possessing the requested affordance, our method also predicts the optimal pose for such functionality and how a potential user can interact with it.
Tuned on only a few synthetic examples across 3 affordance classes, our pipeline achieves a very high success rate on affordance classification and functional pose prediction of 8 classes of novel objects, outperforming learning-based baselines.
Validation through real robot manipulating experiments demonstrates the practical applicability of the imagined user interaction, showcasing the system's ability to independently conceptualize unseen affordances and interact with new objects and scenarios in everyday settings.
%In this study, we examine the issue of checking affordances for new objects and investigate a methodology for interacting with the object through physical means. 
%Our approach involves the utilization of large language models (LLMs) within a framework to automatically analyze novel objects and engage in physical interactions within a simulation environment, with the outcomes being applicable to real-world robot manipulation. 

%The framework comprises three main components: the analysis of affordance imagination, the generation of imagination profiles and the evaluation of imagination results, each containing one or multiple modules driven by LLMs serving distinct functions. 
%To streamline the process and reduce the need for human intervention, the entire workflow, from analyzing novel objects to evaluating outcomes, is carried out autonomously by LLMs. 
%Users are only required to input requested affordance along with basic object information, significantly enhancing process efficiency and reducing human heuristic design. 
%We showcase our methodology in affordance classification using both synthetic models and real objects reconstructed by 3D scanning. 

%Our findings indicate that the framework not only infers its affordance functionality of a novel object but also identifies its functional pose and interaction paradigm. 
%Furthermore, by tuning the prompts for LLMs based on a few data samples, it can be extended to verify various classes of objects by generating tailored simulation profiles.

\end{abstract}

%%%%%%%%%%%%%%%%%%%%%%%%%%%%%%%%%%%%%%%%%%%%%%%%%%%%%%%%%%%%%%%%%%%%%%%%%%%%%%%%

\section{Introduction}

In domestic settings and healthcare facilities, unstructured environments and dynamic daily tasks necessitate human-level intelligence for robots to automatically and adaptively engage in novel objects and scenarios.
Recently, Large Language Models (LLMs) have showcased their impressive conversational and logical abilities. 
Trained on vast amounts of data, LLMs can distill essential information from ambiguous and unspecific requests to create coherent narratives. % with minimal deviations. 
Recent studies have demonstrated that LLMs can assist robots in making high-level decisions that are applicable to real-world scenarios. 
A key challenge lies in the fact that LLMs lack a practical grasp of physics, which hinders their comprehension of the physical world and their ability to make grounded assessments and feasible plans. 
For example, when provided with a description and asked “Is it a table?”, LLMs may provide a logical response that lacks physical plausibility.

Integrating physical properties, the concept of robot imagination assesses the affordances of objects from an interactive perspective, enriching the information for robot manipulation. 
To approach the affordance reasoning from a user-centric perspective, we define the affordance of an object by \textit{Interaction-Based Definition (IBD)}. 
Illustrated by the examples of chairs and containers in \cite{wu2020can, wu2020chair, meng2023prepare}, IBD defines the potential user and feasible user-object interactions, providing extensive instructions for the robot to imagine the scenario and assess the resultant configuration. 
We define the pose of the object that allows the expected interaction as the \textit{functional pose} for the target affordance. 
The object that has at least one functional pose is recognized as functional, \textit{i.e.}, fulfilling such affordance.

\begin{figure}[t]
\centering
\includegraphics[width=0.48\textwidth]{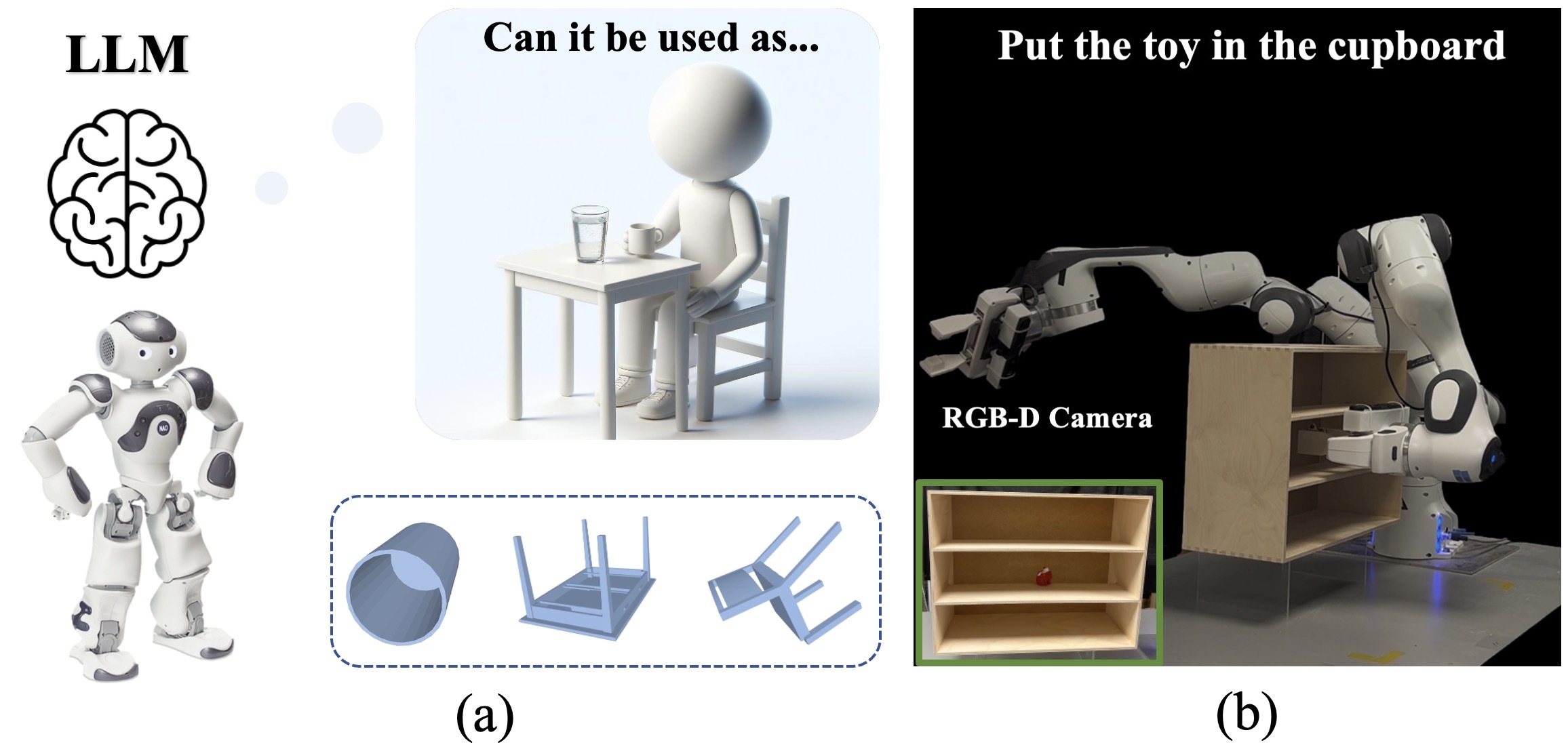}
\caption{Overview of robot affordance imagination with LLMs. (a) The robot imagines the affordances of randomly placed novel objects assisted with LLMs. (b) The robot performs novel tasks based on affordance reasoning.}
\label{Fig. 1}
\vspace{-0.5cm}
\end{figure}

\begin{figure*}[t]
\centering
\includegraphics[width=1.0\textwidth]{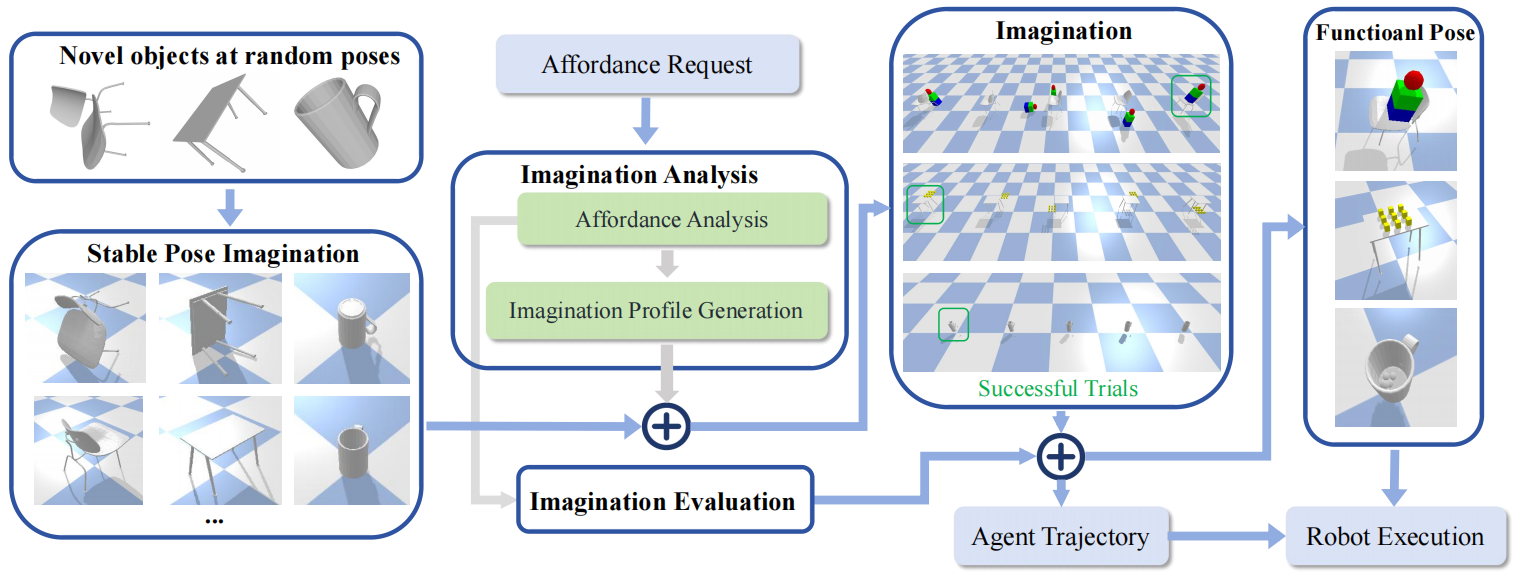}
\caption{Pipeline. Given an object model in a random pose, the algorithm first imagines its stable poses. The Imagination Analyzer analyzes the requested affordance and generates an executable imagination profile. The algorithm simulates the imagination profile with the object and loops for all stable poses. The Imagination Evaluator determines whether the object has the requested affordance. If the object is functional, the functional pose and agent trajectories are recorded for potential real robot execution.}
\label{Fig. 2}
\vspace{-0.5cm}
\end{figure*}

Current robot imagination methods require the development of customized imagination systems for different affordances, in which developers analyze the definition of a class of objects from the dictionary and decompose it into an IBD. 
To translate the affordance request into the programming language that the robot can understand and execute, in previous works \cite{wu2020chair, meng2023prepare}, the simplified agent model is proposed to describe the potential user and encode the interaction into feasible motions and expected outcomes. 
We define the applicable agent models and motions as the \textit{imagination profile}. 
An evaluation matrix is tuned to determine the successful interactions, allowing the robot to recognize the functional poses of the object and, therefore, classify the objects. 
However, it is still an open challenge to imagine novel affordances without complicated implementations by human developers.

In this paper, we tackle this challenge by developing an automatic imagination pipeline that is only conditioned on the name of the requested object affordance, by employing LLMs to replace human analysis and heuristics. 
When approaching novel affordances, humans read the definition from the dictionary, analyze IBD, construct an imagination profile, and put it into the brain to imagine it, thereby analyzing the conclusion and generating executable actions. 
With robots having cameras as eyes, imagination as brains, and end-effectors as hands, we use LLMs as a powerful dictionary that provides detailed profiles.

Instead of asking LLMs to directly reason about the environment and plan actions to manipulate the object, we only require them to answer affordance-related semantic questions that are not conditioned on the specific objects and environments. 
With the imagination profile generated, the robot puts the real object in its brain and imagines the interactions proposed by LLMs (Fig. \ref{Fig. 1}). 
To reach a physically applicable conclusion, LLMs also assist in comparing imagination outcomes with expected ones, providing practical proposals for affordance classification, functional pose prediction, and manipulation.

Evaluated with 301 novel synthetic data, our method showcases a robust 88.2\% accuracy in identifying novel affordances and an impressive 92.7\% success rate in determining functional poses.
In real robot experiments, the system recognizes both the affordances and the functional poses of 18 previously unseen objects. 
It achieves a 100\% success rate in executing novel tasks by accurately parsing semantic requests and reasoning novel affordances. 
Comparing it with leading learning-based approaches and an ablation study baseline, we empathize effectiveness, generality, and practical applicability of our method.
The main contribution of this paper lies in:
\begin{itemize}
    \item An affordance reasoning pipeline that only requests target affordance names.
    \item An imagination framework that simulates customized profiles for multi-class affordances.
    \item A real robot manipulation system for performing novel tasks on unseen objects based on affordance reasoning.
\end{itemize}

\begin{figure*}[t]
\centering
\includegraphics[width=0.9\textwidth]{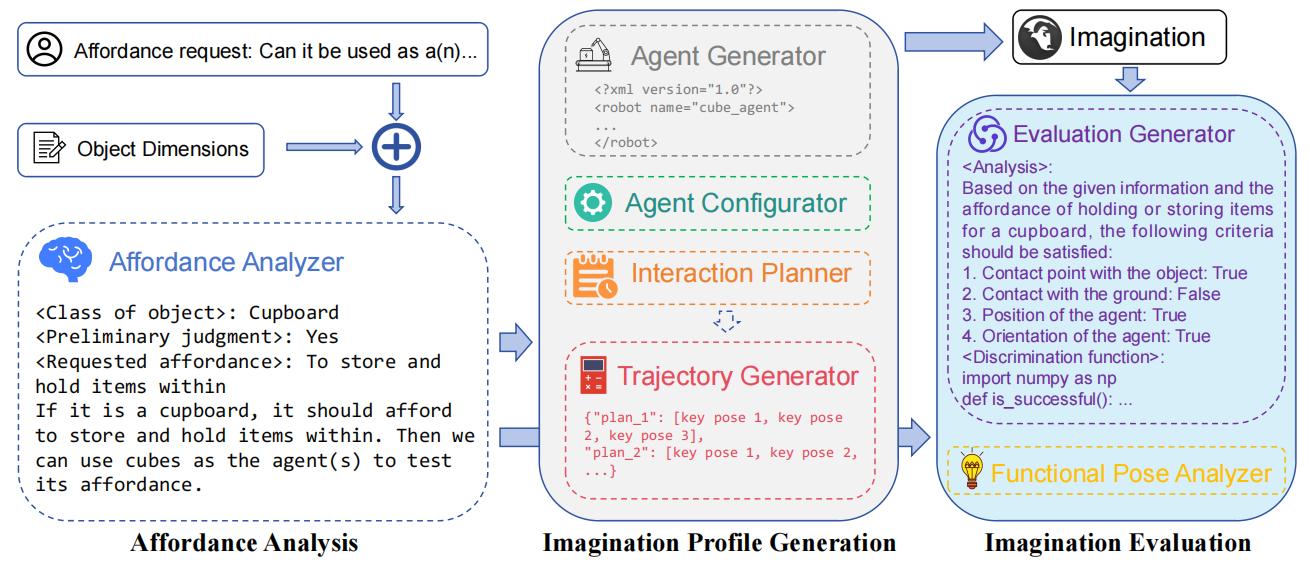}
\caption{Imagination analysis and evaluation framework. The Affordance Analyzer creates the IBD and an abstract imagination outline. The Imagination Profile Generator then develops detailed agent model and action trajectories. Subsequently, the Affordance Evaluator uses a scoring function generated to assess each imagined plan, determining the functional pose.}
\label{Fig. 3}
\vspace{-0.5cm}
\end{figure*}

\section{Related Work}

\subsection{Object Classification}

Object classification is a critical area in computer vision, with Convolutional Neural Networks (CNNs) \cite{lecun1998gradient} forming the foundational approach. Subsequent improvements have significantly enhanced performance on large datasets \cite{krizhevsky2012imagenet,deng2009imagenet,simonyan2014very,he2016deep}. 
Despite their speed and efficacy, these methods impose strict requirements on images. Visual occlusions can significantly impact accuracy, and unnoticed flaws may result in objects lacking their intended functionality (e.g., a cup without a bottom cannot hold liquid).
To address these issues, researchers have integrated multiple viewpoints for 3D object classification \cite{ullman1998three,qi2016volumetric,kanezaki2018rotationnet}. However, these techniques struggle with object poses, leading to reduced classification accuracy when objects are not in their functional poses.

In contrast, instead of relying on visual cues, our approach infers the object's affordance by simulating potential physical interaction with the object in different poses. 
By employing a pre-trained frozen LLM, our system also releases the need for training processes and data requirements.

\subsection{Physical Reasoning of Object Affordance}

Affordance detection has recently become a popular research subject \cite{yu2015fill,hinkle2013predicting,jia20133d,ruiz2018can,ho2023good}, aiding robots in identifying objects, understanding their functionalities, and interacting with them appropriately. 
One approach involves physical interaction, mirroring the intuitive behavior of humans \cite{battaglia2013simulation}. 
Studies \cite{zhu2015understanding, tee2018towards} suggest that robots can explore new object functionalities through interaction, potentially using them as tools for previously inaccessible tasks. Previous research has focused on envisioning affordances by imagination, such as the sittability of a chair \cite{wu2020chair} and the containability of a cup \cite{wu2020can}, and utilizing these functional poses for robot manipulation tasks \cite{meng2023prepare, wu2020can}. 
However, developing these imagination pipelines and evaluation matrices requires extensive effort, and the systems can only identify specific affordances. 
This study streamlines the analysis and evaluation process by leveraging the reasoning capability of LLMs, enabling application to various novel affordances.

\subsection{Robot with Large Language Models}
Vision-Language Models (VLMs) and LLMs have proven effective in broad robotics applications in household, industrial \cite{fan2024embodied}, and laboratory settings \cite{yoshikawa2023large}. 
A significant area of application is language-conditioned object manipulation, where LLMs and VLMs facilitate robot manipulation by aiding in dataset preparation, task analysis, agent policy formulation, and reward function design.
One major challenge in robot learning is the collection and scaling of training data. 
Katara et al. \cite{katara2023gen2sim, xian2023towards} automate the generation of 3D object assets and tasks for simulation environments. 
LLMs define object physics parameters, meaningful tasks, and subtask details. 

LLMs also serve as motion planners \cite{brohan2023rt, liang2024learning, rana2023sayplan, chen2024prompt, stone2023open}, though grounding language commands to low-level actions remains challenging. High-level task planners, such as SayCan \cite{Irpan2022saycan}, combine skill usefulness probabilities and execution success, but often require manual engineering or prior learning.
Some researchers use LLMs to evaluate planning outcomes \cite{yu2023language, triantafyllidis2023intrinsic, kwon2023reward, ma2023eureka}, while hybrid approaches \cite{chen2023autotamp} leverage LLMs for task analysis and traditional planners for motion execution.
Our focus is on leveraging LLMs for affordance reasoning and applying planned motions to manipulation without prior knowledge or manual engineering.

%As interest in large language models grows, the reasoning capability of these models has provided new opportunities for robot planning, a problem that previously required complex algorithm development \cite{ahn2022can,brohan2023rt,chen2023autotamp}. Furthermore, their language analysis skills can help robots better understand user needs based on abstract instructions \cite{ren2023robots,singh2023progprompt,liang2023code}. A unified model suggested in \cite{brohan2023rt} can be implemented across various robots for diverse task executions. Fan et al. highlight the potential of LLMs in industrial robot applications in \cite{fan2024embodied}, with the aim of autonomous completion of production tasks on different assembly lines. One work similar to ours is \cite{xu2023creative}, where a tool recognition system for robot task completion is introduced. It comprises multiple LLM modules including an analyzer, planner, encoder, etc. In contrast to these instances, our focus is on utilizing LLM modules to create an autonomous system that classifies objects by imagining the affordances of novel objects with minimal human intervention.

\section{Language Model Assisted Imagination}

The pose of a rigid body can be represented as $g=(R, \mathbf{p}) \in SE(3)$, where $R \in SO(3)$ is a rotation matrix, $\mathbf{p} \in \mathbb{R}^{3}$ describes the position.
\textit{Functional pose} $g_{f}$ is therefore defined as a pose in which the object affords the functionality.
% We classify the object as functional if there exists at least one functional pose for a requested affordance.
Given an unseen object and a novel affordance, our goal is three-fold:
verifying whether the object possesses such affordance, identifying the functional pose, and determining how to use the object if it is functional.

In this work, we approach the problem of affordance reasoning from an interaction-based perspective by employing a generalized automatic robot imagination pipeline shown in Fig. \ref{Fig. 2}. 
The novel affordance can be assessed by a three-step stream:
1) imagination analysis: analyzing the IBD of the affordance and decomposing it into the potential interaction of agents and expected outcome;
2) imagination: simulating the generated interaction in a physics-based simulator;
3) imagination evaluation: evaluating the quantitative imagination results and proposing the object affordance.
Instead of employing human developers to analyze the affordance definition, we propose an automatic framework that integrates LLMs for a wide scope of affordances.

Taking natural language $a$ as the prompt from the corpus $\mathcal{V}$, LLMs output the optimal content $a^*$ based on previous tokens $t$:
\begin{equation}
    a^* = \arg\max_{a \in \mathcal{V}} P(a | t),
\end{equation}
our framework employs multiple LLM modules to analyze affordances and imagination outcomes.
Specifically, they take the user's prompt $r$ and constraints $c$  as input and generate output $\mathcal{O}=\mathcal{L}(r,c)$.

For a novel object, the input of the framework is a task description $d_{\text{task}}$ which only includes the requested affordance and the dimensional information of the object, defined as $d_{\text{task}}=\{r_{\text{aff}}, g_{\text{obj}}\}$.
The requested affordance $r_{\text{aff}}$ is specified by natural language, and $g_{\text{obj}}$ includes the dimension of the object's bounding box (OBB) and the object position.

In IBD, an object must be in a stable pose to be considered functional for any affordance.
Therefore, the algorithm first takes the 3D model of the unseen object and finds a set of stable poses by stable pose imagination \cite{wu2020chair}, referred to as $g_s \in G_s$.
Setting the object into each stable pose as the candidate pose, by dividing the imagination analysis into a high-level affordance analysis and a low-level imagination profile generation, our approach proposes the four-step workflow:
1) \textbf{Affordance Analyzer} $\mathcal{A}(d_{\text{task}})\rightarrow{(d_{\text{IBD}}, d_{\text{agent}}, d_{\text{interaction}}, d_{\text{evl}})}$ parses the input $d_{\text{task}}$ and generates the IBD $d_{\text{IBD}}$ and an abstract imagination outline including the description of the agent $d_{\text{agent}}$ and agent action $d_{\text{interaction}}$, and expected outcome $d_{\text{evl}}$.
2) \textbf{Imagination Profile Generator} $\mathcal{G}(d_{\text{agent}}, d_{\text{interaction}})\rightarrow{a, \mathcal{T}}$ takes the analyzed outline as input and outputs executable agent model $a$ and trajectories $\mathcal{T}$ for imagination.
3) \textbf{Imagination} $\mathcal{I}(a, \mathcal{T})\rightarrow\mathcal{R}$ simulates the generated agent motions and saves resultant configurations $\mathcal{R}$.
4) \textbf{Imagination Evaluator} $\mathcal{E}(d_{\text{agent}}, d_{\text{interaction}}, d_{\text{evl}}, \mathcal{R})\rightarrow{j}$ outputs the summarized affordance judgments by generating a scoring function to evaluate the imagination outcomes.
The workflow is shown in Fig. \ref{Fig. 3}.

\begin{figure}[t]
\centering
\includegraphics[width=0.5\textwidth]{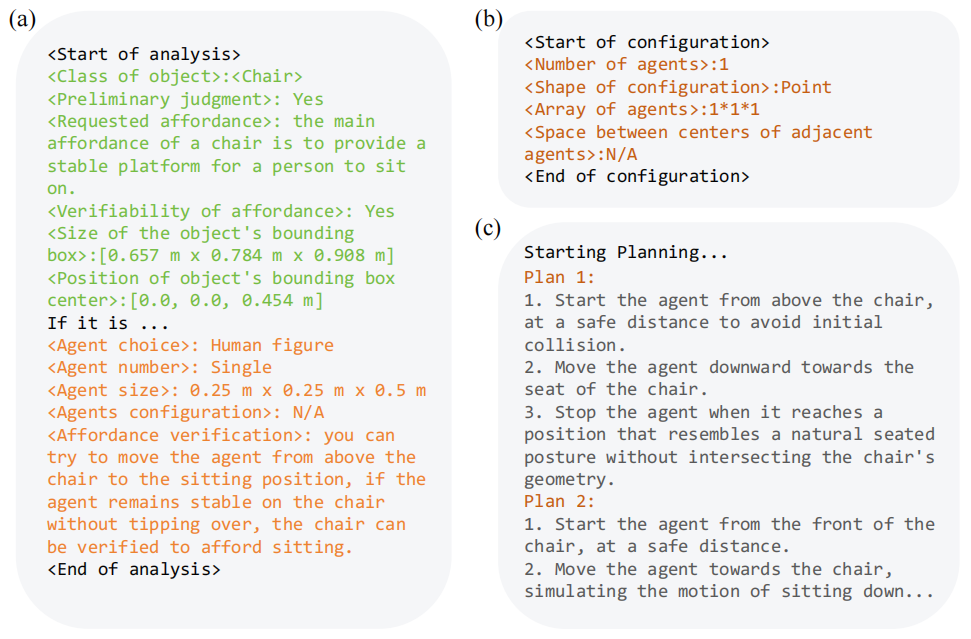}
\caption{(a) Affordance Analyzer, (b) Agent Configuration Generator, (c) Agent Motion Planner.}
%it outputs an outline for the affordance imagination including the agent choice and affordance verification method. (b) Agent Configuration Generator: this is applicable only when multiple agents are used and it generates the number and spatial layout of the agents. (c) Agent Motion Planner: this module generates five motion plans for the agents to interact with objects to imagine their affordances.}
\label{Fig. 4}
\vspace{-0.5cm}
\end{figure}

\subsection{Affordance Analysis}

The Affordance Analyzer $\mathcal{A}$ processes the task description $d_{\text{task}}$ and proposes the outline of the affordance reasoning. This module first assesses whether the object dimension matches the target affordance $r_{\text{aff}}$.
If there is a mismatch, it proposes a potential alternative affordance and replaces it as $r_{\text{aff}}$.
Given the target affordance, the analyzer summarizes the IBD $d_{\text{IBD}}$. %, after which $d_o$ and IBD formulate the task specification, giving an abstract overview for each LLM module.
Based on $d_{\text{IBD}}$, it proposes a physical figure that can act as an agent to verify the target affordance.
%To overcome the limitation of the restricted simulation types, the algorithm simplifies the agents into rigid geometry shapes.
Multiple agents are involved in simulating complicated physics or efficient parallel computing.
The configuration of the agent $d_{\text{agent}}$ is analyzed as a language description of the shape and the abstract layout, \textit{i.e.}, the geometry shape of the agent, and how the agent(s) distribute.
The analyzer plans the active motion $d_{\text{interaction}}$ of the agent(s) to interact with the object and forecast the outcome, $d_{\text{result}}$.
If the resultant configuration of the planned agent(s) motion aligns with the expected outcome, the object is identified as functional.

%Once these conditions are met, the system makes an affordance analysis on the object: "If it is a ..., it should afford ..." and selects another object based on the analysis as the agent to perform physical interaction with the testing object. After choosing the appropriate candidate agent, the module determines that one or multiple agents are needed and if multiple agents are needed then it arranges a rough spatial layout for the agents.

%Following this, the module creates initial interaction method and forecasts for the simulation results. If the outcomes satisfy specific criteria, it is considered that the object possesses the relevant affordance of the specified class. Since the analysis produced in this section is passed as an outline to other LLM modules in later stages, it significantly influences the simulation performance. When designing the output limits, we choose three vase data samples to tune the constraints, making sure that the generated analysis is justified.

\subsection{Imagination Profile Generation}
With the abstract analysis, the algorithm creates a set of configuration files to perform the planned agent motion, including the agent model and the numerical trajectories. The generator is composed of four LLM modules, acting in a step-wise manner. 
%The purpose of this section is to create a set of simulation setup files, encompassing the agent model for interactions, agent configurations, high-level interaction plans, and numerical trajectory parameters. The subsequent text sequentially outlines the roles of these LLM modules.

\subsubsection{Agent Model Generator}
Taking $d_{\text{task}}$ and $d_{\text{agent}}$, the generator outputs a simple representation of the agent model $a$ in a unified robotics description format (URDF) file.
The agent model maintains the essential characteristics related to the target affordance and employs an appropriate scale considering the object dimension.

%Since our framework does not rely on any model assets except the testing model, it generates the agent for interaction on-the-fly. 
%Leveraging the multi-modal output feature of LLM, our system is capable of generating a simplified representation of the agent discussed in the imagination analysis and saving it in a Unified Robot Description Format (URDF) file. 
%For instance, a small sphere may substitute liquid droplets in the imagination of a cup, or a cube replaces the pelvis of a human in the imagination of a chair. 
%Despite the simplification of the model compared to the original analysis, it maintains the essential characteristics, taking into account the appropriate size of the agent relative to the object. Therefore, this factor minimally affects the accuracy of the results.

\subsubsection{Agent Configuration Generator}
To handle complicated scenarios, the imagination analyzer proposes multiple agents to simulate joint or parallel physics.
We organize the spatial distribution of the agent into a cube or planar grid pattern, with sides of the shape aligning with the world frames, and the orientations of all agents remaining the same. 
The distribution is parameterized by the number of agents along each side and the distance between each pair of neighboring agents. 
Based on $d_{\text{task}}$ and $a$, the LLM module proposes the distribution parameters, considering the object dimension and collision avoidance. 
Therefore, the pose of each agent relative to the geometric center of the distribution can be extracted.
For cases where only one agent is employed, the agent geometric center is the center of the distribution.
An example of agent configuration is shown in Fig. \ref{Fig. 4}(b), where only one agent is used.
%This module plays a crucial role when dealing with scenarios involving multiple agents and is primarily responsible for organizing the spatial positioning of these agents. This is necessary as the arrangement needs to be adjusted according to the geometries of different test objects, such as being organized into a 2*2*2 cube or a 3*3 grid pattern. For simplicity, the system does not provide the absolute spatial position of each agent but instead gives its position relative to the geometric center of the array. By employing simple coding and calculations, we can determine the agent's coordinates in the global frame. And it is essential to maintain a reasonable spacing between agents to prevent collisions, a factor taken into account during the module's design to ensure the accuracy of the LLM output.

\subsubsection{Agent Motion Planner}
Taking $d_{\text{task}}$ and the abstract description of the interaction method $d_{\text{interaction}}$, the planner aims to produce high-level plans of agent motions, shown in Fig. \ref{Fig. 4}(c).
Specifically, each plan is given by a sequence of actions $d_{\text{motion}}$, indicating the relative spatial relations between the agent and object, as well as the moving direction of the agent.
With the object placed in each candidate pose, it generates multiple plans $\mathcal{D} = \{d_{\text{motion}1}, d_{\text{motion}2},...,d_{\text{motion}i},...\}$, enabling the exploration of a diverse range of interactions.
$d_{\text{motion}i}$ is the language description of the i-th plan.
%In the initial section, the Imagination Analyzer has put forward sensible recommendations for the interaction method in simulation. The aim of this module is to give a more concrete illustration of the analysis. For instance, determining whether moving from the top implies towards the center of the testing object or deviating in a particular direction, or whether moving towards the object indicates the starting point from a specific direction of the object. These issues impact the success of the experiment (e.g., pouring water into a cup can only be done from above and placing an object in a cupboard can only be done sideways). Consequently, for each potential poses of the object, this module generates five possible plans $\mathcal{X}$ from various directions based on the analysis, primarily focusing on the front, back, left, right and above, and outlines each interaction plan $x_i\in\mathcal{X}$ in a step-by-step manner from the initial to the final point.

\subsubsection{Trajectory Generator}
With the initial language descriptions of the plans, $\mathcal{D}$, the generator converts them into a group of numerical trajectories, referred to as $\mathcal{T}=\{t_1,t_2,...,t_i,...\}$.
$t_i$ is the i-th trajectory of the center of the agent(s).
Each trajectory is given as a sequence of via poses of the agent(s) relative to the object, with the object in a candidate pose.
The algorithm calculates the trajectory of each agent as the executable plan saved in a JSON file, which can be imported for imagination.

%After generating feasible plans in the preceding stage, this LLM module converts the abstract description in $\mathcal{X}$ into numerical parameters to import into the simulator, which refers to the motion trajectories $\mathcal{T}=\{t_1,t_2,...,t_5\}$ of the agent (or the trajectory of the center of the layout in the case of multiple agents), where $t_i$ represents each individual plan. According to multiple steps in the plan description, the trajectory can be divided into two or three key poses, and these poses are connected to form the trajectory of this plan. Since the module considers the agent and the size of the object, the test shows that the collision between the two in the trajectory is effectively avoided.

\subsection{Imagination}
The algorithm imagines the proposed agent $a$ and planned interactions $\mathcal{T}$ in a physics-based simulator.
It first loads the objects in the candidate stable poses and the agents in the initial poses, then simulates the agent moving through the planned via-poses.
The agent is released after the trajectory is completely executed. 
In addition, collision checking is performed at each step, and when a collision occurs, all agents involved are released immediately.

For the simulation of each plan, the resultant configuration $r$ is summarized from two aspects:
1) Agent poses: the position and orientation of each agent relative to the object;
2) Contact points: number of contact points of each agent with the object, other agents, and the ground, respectively.

The outcome of imagination is the combination of the resultant configuration of all generated plans $\mathcal{R} = \{r_1,r_2,...,r_i,...\}$.

\subsection{Affordance Evaluation}
In this part, the algorithm analyzes the results of the simulation $\mathcal{R}$ to evaluate the object affordance.
It consists of two LLM modules to evaluate $r$ of each plan and the overall object affordance, respectively.

%for determining the success of a single trial (each interaction plan), which implies whether the affordance is verified, and for combining the results of all the plans in order to determine whether the testing object is in the functional pose.
\subsubsection{Scoring Function Generator}
Based on $d_{\text{IBD}}$ and $d_{\text{evl}}$, the generator produces an executable function $\mathcal{F}$ that scores the outcome of each plan.
The function takes in the resultant configuration of each imagined plan $r$ and generates a weighted score $S$.
\begin{equation}
    \mathcal{F} (r) \longrightarrow S
\end{equation}
It considers criteria related to target affordance, including agent-object relative position, agent orientation, agent-object contact, agent-agent contact, and agent-ground contact.
The success of the plan is determined by whether $S$ exceeds 0.
All successful interactions are selected as candidate functional interactions $(d_{\text{motion}f}, t_f, r_f)$.
%which may be defined as the proportion of successful agents in a multi-agent scenario or as 0 in the case of a single agent.
Compared to previous works where the evaluation matrix is manually defined and tuned, our method provides an automatic approach that enables generalization across various types of affordances.

\subsubsection{Functional Pose Analyzer}
The imagined pose is hypothesized as potentially functional if there exists at least one candidate functional interaction.
To analyze the validity of the candidate functional pose, we introduce another LLM module.
By analyzing the candidate functional interactions $(d_{\text{motion}f}, t_f, r_f)$, it determines if the evaluation provided by $\mathcal{F}$ is consistent with common sense.

The object is classified as not functional if none of the candidate interactions of all stable poses $g_s \in G_s$ is valid.
If there exists at least one valid candidate interaction, the analyzer selects the best functional interaction.
The corresponding object pose is therefore the optimal functional pose.

%With all stable poses $G_s$ as candidates, the algorithm examines the $S$ of every interaction plan in each $g_s \in G_s$ and the object affords the target affordance if it finds at least one pose where an interaction plan is considered successful. 
%And to confirm whether it is a functional pose, we introduce an additional LLM module to assess the interaction plans and their outcomes in this pose and determines whether the success of a certain plan is consistent with common sense. 
%This module further improves the accuracy of the system by leveraging the reasoning capability of LLMs to determine the function pose only when the affordance is found and successful plans in this pose are justified.
% The imagined agent resultant poses in a successful plan when the object is placed in a functional pose are recognized as agent functional pose.

\begin{table*}[t]
\caption{Accuracy (\%) of Affordance Classification}
\label{Tab. 1}
\renewcommand\arraystretch{1.5}
\tabcolsep=0.25cm
\centering
\begin{tabular}{ccccccccccc}
\Xhline{2\arrayrulewidth}
\multicolumn{2}{c}{\multirow{2}{*}{Method}} & \multicolumn{9}{c}{Synthetic data}                                \\ \cline{3-11}
\multicolumn{2}{c}{}                        & basket & bathtub & chair & cup & plate & table & vase & bowl & overall                           \\ \hline
\multicolumn{2}{c}{\textbf{Ours} (random pose)}                    & \textbf{94.6}   & \textbf{82.6}       & 85.7       & \textbf{91.2}      & \textbf{91.7}      & \textbf{85.7}     & 74.6  & \textbf{93.2}   & \textbf{87.7}                                                    \\ \hline
\multirow{2}{*}{BLIP \cite{li2022blip}}   & functional pose   & 45.2    & 13.0       & 95.2        & 70.6      & 11.1      & 85.1      & 87.2     & 68.2    & 56.8                            \\ \cline{2-11} 
                        & random pose       & 12.9    & 8.7       & 42.9        & 35.3      & 5.6      & 31.9      & 38.5     & 36.4         & 30.2                         \\ \Xhline{2\arrayrulewidth}
\end{tabular}
\end{table*}

\section{Experiments}

\begin{figure}[t!]
\centering
\includegraphics[width=0.40\textwidth]{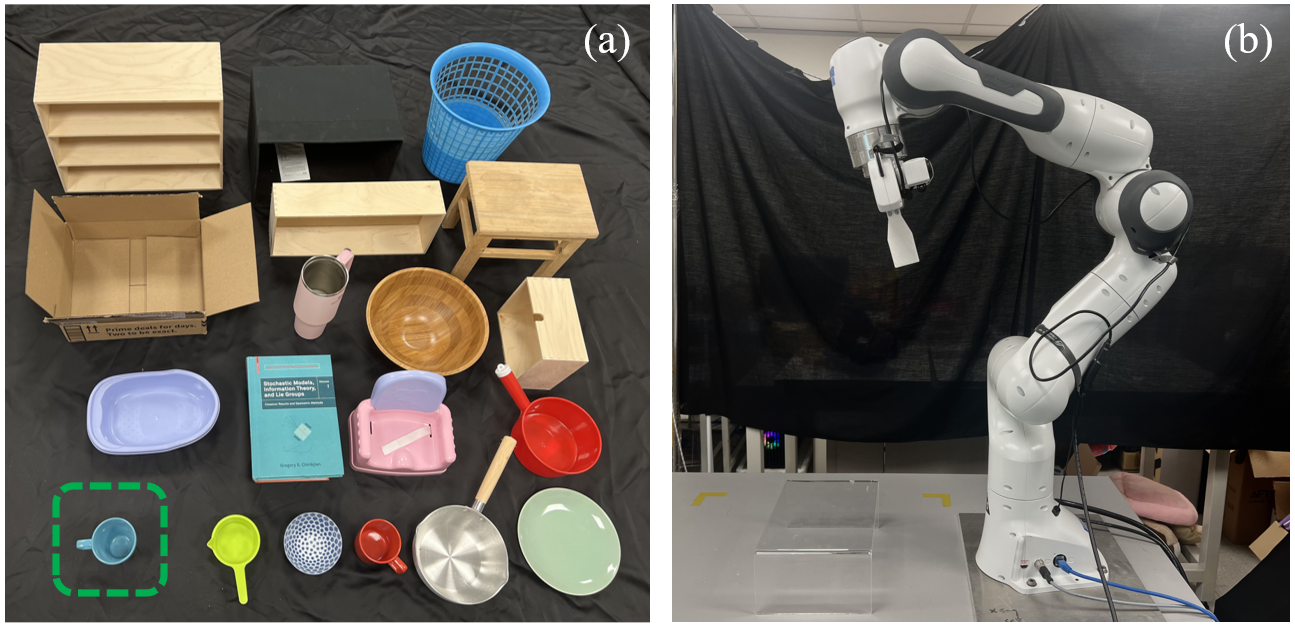}
\caption{Real world experiment details. (a) Snapshot for different classes of objects used for affordance imagination. The circled cup is used to tune the robot planning. (b) Real robot setting.}
\label{Fig. 5}
\vspace{-0.5cm}
\end{figure}

We implement our framework system with Python, using Pybullet \cite{coumans2016pybullet} as the physics simulator. 
The algorithm is evaluated on a computer running an Intel Core i7-11370H @ 3.3GHz CPU and Nvidia GTX 3060 GPU. 
In a real experiment setting Fig. \ref{Fig. 5} (b), a Franka Emika robot arm is used for manipulation. 
An RGBD camera is mounted on the end effector for reconstruction. 
The language model in this study is founded on GPT-4 \cite{achiam2023gpt}. 
To enhance response time and speed of generation, we opt for the GPT-4-turbo variant and set the parameter \textit{Temperature = 0.1} to ensure optimality while providing a certain degree of generalization.

\subsection{Data}
Our dataset comprises 301 synthetic and 19 real-life objects. 
The synthetic objects, sourced from a subset of the Princeton ModelNet40 dataset \cite{wu20153d}, span 8 classes: cup, basket, bathtub, chair, plate, table, vase, and bowl.
To optimize the performance of LLM modules and the simulator, we utilize 6 synthetic objects across 3 classes, \textit{i.e.}, cup, vase, and chair, to tune the prompt.
The rest data, which are unseen by our system, are used as the test set.
To assess affordance classification, non-functional objects from various classes are also included in the testing phase.
%To test each category of affordance, we also randomly select objects from other classes as non-functional data.

In real robot experiments, we use a cup to tune the parameters for the robot planning to shorten the sim-to-real gap.
The remaining 18 novel objects, varied in size and appearance, are used to test the recognition of 15 classes of affordances and affordance-based manipulation.
This includes 13 novel affordances previously unencountered by our system.

\subsection{Real Robot Experiment}
The object is placed on a transparent stand to enable comprehensive scanning of the bottom section, with it in a functional pose.
The task is given to the system through a semantic request formatted as “put the (real agent) in/on the (requested affordance name).”
The system has no prior knowledge of the object as well as the requested affordance.

The robot arm first moves to 12 predefined poses to scan the object.
The object model is cropped and segmented from the scene point cloud reconstructed using TSDF-Fusion \cite{curless1996volumetric}.
With the requested affordance name segmented and passed to the affordance reasoning module, the algorithm imagines the object model and classifies if the object can be used with the requested affordance in the placed pose.
If the object is recognized as functional and in a functional pose, the algorithm proposes an optimal agent trajectory.
The robot then requests a volunteer to hand over the real agent.
Having the real agent in hand, the robot executes the imagined trajectory and positions the agent, transferring the affordance understanding to practical actions.
Throughout this process, simple force control is used for robot manipulation, pausing its motion if external force surpasses a pre-defined limit, ensuring safety and precision in task execution.

\subsection{Baseline}

We compare our method with the baseline, BLIP \cite{li2022blip}, which is a unified vision-language model for image understanding and generation, and we use the version fined-tuned on Visual Question Answering (VQA) task that makes responses to user's prompt, we make comparisons on two criteria: affordance classification and functional pose prediction.

For affordance classification, we present an object's image alongside the query “Is it a (requested affordance name)?”, and it ascertains whether the model possesses the requested affordance. 
In functional pose prediction, we input the image together with the query "Can the (requested affordance name) function in this pose?"
To study how the object's pose impacts its performance and demonstrate the versatility of our approach across different poses, for each data sample we capture two images in arbitrary and functional poses, respectively.  

Additionally, we perform ablation experiments in functional pose prediction to assess the significance of the functional pose analyzer in the framework. 
By comparing the outcomes of the base method with and without the functional pose analyzer, we aim to elucidate its impact on the overall performance of the affordance reasoning process.

\begin{figure}[t]
\centering
\includegraphics[width=0.48\textwidth]{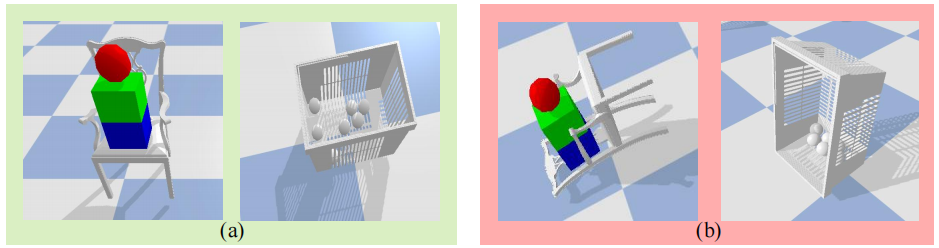}
\caption{Functional poses analysis. (a) True functional poses. (b) "Fake" functional poses in which the simulation result is scored as successful but is not validated by the Functional Pose Analyzer.}
\label{Fig. 6}
\vspace{-0.5cm}
\end{figure}

\subsection{Evaluation}
%For the results of each test sample, we use two evaluation criteria: affordance classification and functional pose prediction. 
%For the imagination of affordance, the judgment is given by the scoring function, and the affordance is verified if the simulation results of the agent satisfy all of the conditions taken into considerations and in the case of multiple agents the proportion of agents satisfying the conditions exceeds a threshold. 
%For the functional pose prediction, for each object for which affordance is verified the system stores its convinced functional pose in which simulation result shows success as the functional pose, and if there are more than one success trials, the one with the highest score is selected. 
We recruit volunteers to annotate the experiment result.
For each object, we present it to the volunteer and ask “Do you think it can be a (requested affordance name)?”
For each predicted functional pose, with the result of imagination, we ask the volunteer “Do you think this is the functional pose for (requested affordance name)?”
In addition, for each trial of the real robot experiment, we show the experiment video and ask “Do you think the task was successfully performed?” to ensure that the pipeline runs in a reasonable manner.
%The saved functional pose is then judged by our volunteer by asking "Do you think this is the functional pose for..." to check whether it is consistent with common sense when it provides functionality.

\begin{figure*}[t!]
\centering
\includegraphics[width=1.0\textwidth]{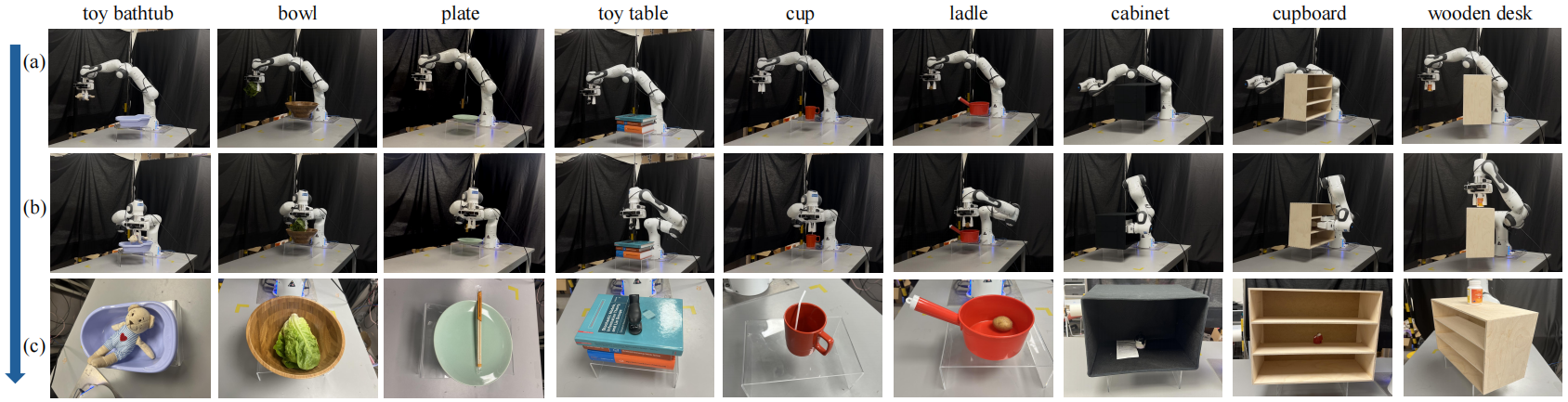}
\caption{Real Robot Experiment Results. (a)-(b)The robot positions the real agent according to the requested task. (c) Results.}
\label{Fig. 7}
\vspace{-0.5cm}
\end{figure*}

\section{Results}

We test our method on synthetic and real object data, respectively.
The synthetic data are placed in a random pose, while the real object is placed on the table with an arbitrary upright pose.
%and from the logs it is shown that without additional optimization our pipeline takes about 4 minutes to generate the profiles used for the simulation, and about 15s on imagination of each stable pose for an object model.

\subsection{Affordance Classification}

The affordance classification achieves high success rates on synthetic data, as shown in Tab. \ref{Tab. 1}.
Notably, the classification of novel affordance classes demonstrates exceptional performance, achieving an 88.2\% success rate.
Failure examples are mainly due to improperly generated profiles, such as imagination outline analysis, agent's model, trajectory parameters of agent motion, etc.
We notice that the performance is degraded on vases, which we hypothesize is because the geometry varies greatly and the openings are usually small. 
In the absence of specific information about geometry, the generated agent may only match the overall dimensions of the object and it might be oversized for the opening, leading to a failure of the affordance imagination.
BLIP does not perform as well on synthetic data even with objects in a functional pose.
It shows good performances only for classes with unique appearance features such as chairs and vases.
When objects are placed in random poses, it results in an overall correctness of only 30.2\%. 

In real robot experiments, the success rate is achieved 100\%, with the objects placed in functional poses. 
BLIP showed a significant increase in performance on real data, with a success rate of 84.2\%, which we attribute to the fact that most of the data used to train the model come from pictures of real objects.

% \begin{table}[t!]
% \caption{Accuracy (\%) of Functional Pose Prediction}
% \label{Tab. 2}
% \renewcommand\arraystretch{1.5}
% \tabcolsep=0.1cm
% \centering
% \begin{tabular}{ccc}
% \Xhline{2\arrayrulewidth}
% Method                              & Synthetic data & Real data \\ \hline
% Base                                & 92.7          & 91.7      \\ 
% Base w/o Functional Pose Analyzer & 75.3          & 79.2      \\ \Xhline{2\arrayrulewidth}
% \end{tabular}
% \end{table}

\begin{table}[ht!]
\caption{Accuracy (\%) of Functional Pose Prediction}
\label{Tab. 2}
\renewcommand\arraystretch{1.5}
\tabcolsep=0.2cm
\centering
\begin{tabular}{ccc}
\Xhline{2\arrayrulewidth}
Method                            & Synthetic data & Real data \\ \hline
\textbf{Ours}                              & \textbf{92.7}           & \textbf{100.0}      \\ \hline
Ours w/o Functional Pose Analyzer & 75.3           & 79.2      
\\ \hline
BLIP \cite{li2022blip} & 55.1           & 60.5 \\
\Xhline{2\arrayrulewidth}
\end{tabular}
\end{table}

\subsection{Functional Pose Prediction}
We evaluate the functional pose prediction results on objects that are successfully recognized as functional.
In Tab. \ref{Tab. 2}, our method achieves very high accuracy in functional pose prediction for synthetic data.
When considering only the current pose, the robot successfully recognizes the object as being in a functional pose across all real-world trials. 
%In addition, an ablation study is conducted by removing the Functional Pose Analyzer, which result exhibits a notable decline in accuracy, suggesting that it plays an important role in accurately determining the correct functional pose. We infer that the cause of the failure may be the incomprehension of the Scoring Function, leading to the identification of “fake” functional poses that we do not expect from human common sense, illustrated in Fig. \ref{Fig. 6}.

In contrast, the baseline method BLIP shows low accuracy on both synthetic and real data, which is in line with the out-of-distribution challenge faced by vision-based methods.
The performance is hugely affected by the viewpoints, object pose, and object appearances.
The ablation trials exhibit a notable decline in accuracy, suggesting that the Functional Pose Analyzer plays an important role in accurately determining the correct functional pose.
Failure is often caused by the incomprehension of the Scoring Function, leading to the identification of “fake” functional poses that we do not expect from human common sense, illustrated in Fig. \ref{Fig. 6} By incorporating the Functional Pose Analyzer, our method provides more adequate and reliable judgments, ensuring that the predictions align more closely with practical expectations.

%In contrast, , which is in line with the challenge faced by vision-based methods we mentioned before: They are usually unable to judge whether the object is in a functional pose, and affected by the viewing points, the pose of the object also impacts its classification accuracy.
%After our affordance imagination, for the successful data samples we also explore the correctness of their recorded functional poses, this is because it may be the case that the result yields successful judgment for a particular agent motion plan for a certain pose that is not functional, examples are shown in Fig. \ref{Fig. 6}, which we call "fake" functional poses that we do not expect from human common sense. From the result in Tab. \ref{Tab. 2}, our method achieves a high accuracy of more than 90\% in functional pose prediction for both synthetic and real object data. For comparison, we remove the Functional Pose Analyzer from the framework and run the experiment under the same condition, and the result shows a decrease in accuracy of more than 10\% without this module, suggesting that it plays an important role in the screening of the correct functional pose.

\subsection{Real Robot Manipulation}
Qualitative results are shown in Fig. \ref{Fig. 7}, our system achieves 100\% across 20 novel trials encompassing 15 distinct tasks, utilizing 18 objects previously unknown to the system.
Notably, the system exhibits impressive generalization capabilities by successfully recognizing and manipulating 13 new affordances.
Furthermore, it can identify objects with multi-functional usage and adapts its interaction accordingly.
For example, the rightmost two trials in Fig. \ref{Fig. 7} showcase an object that is recognized as possessing two types of affordances, \textit{i.e.}, a shelf by positioning the agent on top, and a cupboard by inserting the agent inside.

\section{Conclusion}
In this paper, we introduced an intelligent real2sim2real affordance reasoning framework that enables robots to understand and interact with novel classes of objects based on semantic requests.
This process involves analyzing the affordance, imagining the generated scenarios, and evaluating the outcome to classify object affordance, predict the functional pose, and propose the potential user interaction.
%We developed a real robot manipulation system to automatically reconstruct the object, reason the affordance, and use the object intelligently by grounding the reasoned user interaction into robot motion planning.
Our system demonstrated a success rate of 88.2\% in identifying the affordance of novel classes, and successfully performed 20 novel tasks in real-world settings, showing significant potential in a wide range of daily indoor applications.
%The ability to conceptualize novel affordances and act on unseen object instances allows the robot to automatically navigate unknown environments and perform novel requested tasks, such as tidying bedrooms, preparing dining tables, and cleaning kitchens.
Future work aims to expand the framework to articulated and deformable objects and understand and execute more complicated task commands.

%In this paper, a framework is introduced for the autonomous analysis and execution of novel objects affordance imagination. The method not only identifies the class of an object by checking its affordance but also searches for its functional pose and explores how other objects would interact with it to utilize its functionality, which information can guide the robot in manipulating the object in real-world scenarios. The experimental outcomes demonstrate that our approach reduces the need for object orientation as opposed to appearance-based visual classification. Additionally, by leveraging LLM, we can extend the affordance imagination to various object classes. Although the framework currently has some constraints, it offers insight for physical reasoning in this domain.


%%%%%%%%%%%%%%%%%%%%%%%%%%%%%%%%%%%%%%%%%%%%%%%%%%%%%%%%%%%%%%%%%%%%%%%%%%%%%%%%

%%%%%%%%%%%%%%%%%%%%%%%%%%%%%%%%%%%%%%%%%%%%%%%%%%%%%%%%%%%%%%%%%%%%%%%%%%%%%%%%

%%%%%%%%%%%%%%%%%%%%%%%%%%%%%%%%%%%%%%%%%%%%%%%%%%%%%%%%%%%%%%%%%%%%%%%%%%%%%%%%

%%%%%%%%%%%%%%%%%%%%%%%%%%%%%%%%%%%%%%%%%%%%%%%%%%%%%%%%%%%%%%%%%%%%%%%%%%%%%%%%
\bibliographystyle{ieeetr}
\bibliography{ref}

\begin{thebibliography}{10}

\bibitem{wu2020can}
H.~Wu and G.~S. Chirikjian, ``Can i pour into it? robot imagining open containability affordance of previously unseen objects via physical simulations,'' {\em IEEE Robotics and Automation Letters}, vol.~6, no.~1, pp.~271--278, 2020.

\bibitem{wu2020chair}
H.~Wu, D.~Misra, and G.~S. Chirikjian, ``Is that a chair? imagining affordances using simulations of an articulated human body,'' in {\em 2020 IEEE International Conference on Robotics and Automation (ICRA)}, pp.~7240--7246, IEEE, 2020.

\bibitem{meng2023prepare}
X.~Meng, H.~Wu, S.~Ruan, and G.~S. Chirikjian, ``Prepare the chair for the bear! robot imagination of sitting affordance to reorient previously unseen chairs,'' {\em arXiv preprint arXiv:2306.11448}, 2023.

\bibitem{lecun1998gradient}
Y.~LeCun, L.~Bottou, Y.~Bengio, and P.~Haffner, ``Gradient-based learning applied to document recognition,'' {\em Proceedings of the IEEE}, vol.~86, no.~11, pp.~2278--2324, 1998.

\bibitem{krizhevsky2012imagenet}
A.~Krizhevsky, I.~Sutskever, and G.~E. Hinton, ``Imagenet classification with deep convolutional neural networks,'' {\em Advances in neural information processing systems}, vol.~25, 2012.

\bibitem{deng2009imagenet}
J.~Deng, W.~Dong, R.~Socher, L.-J. Li, K.~Li, and L.~Fei-Fei, ``Imagenet: A large-scale hierarchical image database,'' in {\em 2009 IEEE conference on computer vision and pattern recognition}, pp.~248--255, Ieee, 2009.

\bibitem{simonyan2014very}
K.~Simonyan and A.~Zisserman, ``Very deep convolutional networks for large-scale image recognition,'' {\em arXiv preprint arXiv:1409.1556}, 2014.

\bibitem{he2016deep}
K.~He, X.~Zhang, S.~Ren, and J.~Sun, ``Deep residual learning for image recognition,'' in {\em Proceedings of the IEEE conference on computer vision and pattern recognition}, pp.~770--778, 2016.

\bibitem{ullman1998three}
S.~Ullman, ``Three-dimensional object recognition based on the combination of views,'' {\em Cognition}, vol.~67, no.~1-2, pp.~21--44, 1998.

\bibitem{qi2016volumetric}
C.~R. Qi, H.~Su, M.~Nie{\ss}ner, A.~Dai, M.~Yan, and L.~J. Guibas, ``Volumetric and multi-view cnns for object classification on 3d data,'' in {\em Proceedings of the IEEE conference on computer vision and pattern recognition}, pp.~5648--5656, 2016.

\bibitem{kanezaki2018rotationnet}
A.~Kanezaki, Y.~Matsushita, and Y.~Nishida, ``Rotationnet: Joint object categorization and pose estimation using multiviews from unsupervised viewpoints,'' in {\em Proceedings of the IEEE conference on computer vision and pattern recognition}, pp.~5010--5019, 2018.

\bibitem{yu2015fill}
L.-F. Yu, N.~Duncan, and S.-K. Yeung, ``Fill and transfer: A simple physics-based approach for containability reasoning,'' in {\em Proceedings of the IEEE international conference on computer vision}, pp.~711--719, 2015.

\bibitem{hinkle2013predicting}
L.~Hinkle and E.~Olson, ``Predicting object functionality using physical simulations,'' in {\em 2013 IEEE/RSJ International Conference on Intelligent Robots and Systems}, pp.~2784--2790, IEEE, 2013.

\bibitem{jia20133d}
Z.~Jia, A.~Gallagher, A.~Saxena, and T.~Chen, ``3d-based reasoning with blocks, support, and stability,'' in {\em Proceedings of the IEEE Conference on Computer Vision and Pattern Recognition}, pp.~1--8, 2013.

\bibitem{ruiz2018can}
E.~Ruiz and W.~Mayol-Cuevas, ``Where can i do this? geometric affordances from a single example with the interaction tensor,'' in {\em 2018 IEEE International Conference on Robotics and Automation (ICRA)}, pp.~2192--2199, IEEE, 2018.

\bibitem{ho2023good}
S.-B. Ho, ``Why is that a good or not a good frying pan?--knowledge representation for functions of objects and tools for design understanding, improvement, and generation,'' in {\em 2023 IEEE Symposium Series on Computational Intelligence (SSCI)}, pp.~121--128, IEEE, 2023.

\bibitem{battaglia2013simulation}
P.~W. Battaglia, J.~B. Hamrick, and J.~B. Tenenbaum, ``Simulation as an engine of physical scene understanding,'' {\em Proceedings of the National Academy of Sciences}, vol.~110, no.~45, pp.~18327--18332, 2013.

\bibitem{zhu2015understanding}
Y.~Zhu, Y.~Zhao, and S.~Chun~Zhu, ``Understanding tools: Task-oriented object modeling, learning and recognition,'' in {\em Proceedings of the IEEE Conference on Computer Vision and Pattern Recognition}, pp.~2855--2864, 2015.

\bibitem{tee2018towards}
K.~P. Tee, J.~Li, L.~T.~P. Chen, K.~W. Wan, and G.~Ganesh, ``Towards emergence of tool use in robots: Automatic tool recognition and use without prior tool learning,'' in {\em 2018 IEEE International Conference on Robotics and Automation (ICRA)}, pp.~6439--6446, IEEE, 2018.

\bibitem{fan2024embodied}
H.~Fan, X.~Liu, J.~Y.~H. Fuh, W.~F. Lu, and B.~Li, ``Embodied intelligence in manufacturing: leveraging large language models for autonomous industrial robotics,'' {\em Journal of Intelligent Manufacturing}, pp.~1--17, 2024.

\bibitem{yoshikawa2023large}
N.~Yoshikawa, M.~Skreta, K.~Darvish, S.~Arellano-Rubach, Z.~Ji, L.~Bj{\o}rn~Kristensen, A.~Z. Li, Y.~Zhao, H.~Xu, A.~Kuramshin, {\em et~al.}, ``Large language models for chemistry robotics,'' {\em Autonomous Robots}, vol.~47, no.~8, pp.~1057--1086, 2023.

\bibitem{katara2023gen2sim}
P.~Katara, Z.~Xian, and K.~Fragkiadaki, ``Gen2sim: Scaling up robot learning in simulation with generative models,'' {\em arXiv preprint arXiv:2310.18308}, 2023.

\bibitem{xian2023towards}
Z.~Xian, T.~Gervet, Z.~Xu, Y.-L. Qiao, T.-H. Wang, and Y.~Wang, ``Towards generalist robots: A promising paradigm via generative simulation,'' {\em arXiv preprint arXiv:2305.10455}, 2023.

\bibitem{brohan2023rt}
A.~Brohan, N.~Brown, J.~Carbajal, Y.~Chebotar, X.~Chen, K.~Choromanski, T.~Ding, D.~Driess, A.~Dubey, C.~Finn, {\em et~al.}, ``Rt-2: Vision-language-action models transfer web knowledge to robotic control,'' {\em arXiv preprint arXiv:2307.15818}, 2023.

\bibitem{liang2024learning}
J.~Liang, F.~Xia, W.~Yu, A.~Zeng, M.~G. Arenas, M.~Attarian, M.~Bauza, M.~Bennice, A.~Bewley, A.~Dostmohamed, {\em et~al.}, ``Learning to learn faster from human feedback with language model predictive control,'' {\em arXiv preprint arXiv:2402.11450}, 2024.

\bibitem{rana2023sayplan}
K.~Rana, J.~Haviland, S.~Garg, J.~Abou-Chakra, I.~Reid, and N.~Suenderhauf, ``Sayplan: Grounding large language models using 3d scene graphs for scalable robot task planning,'' in {\em 7th Annual Conference on Robot Learning}, 2023.

\bibitem{chen2024prompt}
Y.~Chen, J.~Arkin, Y.~Hao, Y.~Zhang, N.~Roy, and C.~Fan, ``Prompt optimization in multi-step tasks (promst): Integrating human feedback and preference alignment,'' {\em arXiv preprint arXiv:2402.08702}, 2024.

\bibitem{stone2023open}
A.~Stone, T.~Xiao, Y.~Lu, K.~Gopalakrishnan, K.-H. Lee, Q.~Vuong, P.~Wohlhart, S.~Kirmani, B.~Zitkovich, F.~Xia, {\em et~al.}, ``Open-world object manipulation using pre-trained vision-language models,'' {\em arXiv preprint arXiv:2303.00905}, 2023.

\bibitem{Irpan2022saycan}
A.~Irpan, A.~Herzog, A.~T. Toshev, A.~Zeng, A.~Brohan, B.~A. Ichter, B.~David, C.~Parada, C.~Finn, C.~Tan, D.~Reyes, D.~Kalashnikov, E.~V. Jang, F.~Xia, J.~L. Rettinghouse, J.~C. Hsu, J.~L. Quiambao, J.~Ibarz, K.~Rao, K.~Hausman, K.~Gopalakrishnan, K.-H. Lee, K.~A. Jeffrey, L.~Luu, M.~Yan, M.~S. Ahn, N.~Sievers, N.~J. Joshi, N.~Brown, O.~E.~E. Cortes, P.~Xu, P.~P. Sampedro, P.~Sermanet, R.~J. Ruano, R.~C. Julian, S.~A. Jesmonth, S.~Levine, S.~Xu, T.~Xiao, V.~O. Vanhoucke, Y.~Lu, Y.~Chebotar, and Y.~Kuang, ``Do as i can, not as i say: Grounding language in robotic affordances,'' 2022.

\bibitem{yu2023language}
W.~Yu, N.~Gileadi, C.~Fu, S.~Kirmani, K.-H. Lee, M.~G. Arenas, H.-T.~L. Chiang, T.~Erez, L.~Hasenclever, J.~Humplik, {\em et~al.}, ``Language to rewards for robotic skill synthesis,'' {\em arXiv preprint arXiv:2306.08647}, 2023.

\bibitem{triantafyllidis2023intrinsic}
E.~Triantafyllidis, F.~Christianos, and Z.~Li, ``Intrinsic language-guided exploration for complex long-horizon robotic manipulation tasks,'' {\em arXiv preprint arXiv:2309.16347}, 2023.

\bibitem{kwon2023reward}
M.~Kwon, S.~M. Xie, K.~Bullard, and D.~Sadigh, ``Reward design with language models,'' {\em arXiv preprint arXiv:2303.00001}, 2023.

\bibitem{ma2023eureka}
Y.~J. Ma, W.~Liang, G.~Wang, D.-A. Huang, O.~Bastani, D.~Jayaraman, Y.~Zhu, L.~Fan, and A.~Anandkumar, ``Eureka: Human-level reward design via coding large language models,'' {\em arXiv preprint arXiv:2310.12931}, 2023.

\bibitem{chen2023autotamp}
Y.~Chen, J.~Arkin, Y.~Zhang, N.~Roy, and C.~Fan, ``Autotamp: Autoregressive task and motion planning with llms as translators and checkers,'' {\em arXiv preprint arXiv:2306.06531}, 2023.

\bibitem{li2022blip}
J.~Li, D.~Li, C.~Xiong, and S.~Hoi, ``Blip: Bootstrapping language-image pre-training for unified vision-language understanding and generation,'' in {\em International Conference on Machine Learning}, pp.~12888--12900, PMLR, 2022.

\bibitem{coumans2016pybullet}
E.~Coumans and Y.~Bai, ``Pybullet, a python module for physics simulation for games, robotics and machine learning,'' 2016.

\bibitem{achiam2023gpt}
J.~Achiam, S.~Adler, S.~Agarwal, L.~Ahmad, I.~Akkaya, F.~L. Aleman, D.~Almeida, J.~Altenschmidt, S.~Altman, S.~Anadkat, {\em et~al.}, ``Gpt-4 technical report,'' {\em arXiv preprint arXiv:2303.08774}, 2023.

\bibitem{wu20153d}
Z.~Wu, S.~Song, A.~Khosla, F.~Yu, L.~Zhang, X.~Tang, and J.~Xiao, ``3d shapenets: A deep representation for volumetric shapes,'' in {\em Proceedings of the IEEE conference on computer vision and pattern recognition}, pp.~1912--1920, 2015.

\bibitem{curless1996volumetric}
B.~Curless and M.~Levoy, ``A volumetric method for building complex models from range images,'' in {\em Proceedings of the 23rd annual conference on Computer graphics and interactive techniques}, pp.~303--312, 1996.

\end{thebibliography}

\end{document}